# Directional Cross Diamond Search Algorithm for Fast Block Motion Estimation

Hongjun Jia, Li Zhang

Tsinghua University, Beijing, China

Abstract—In block-matching motion estimation (BMME), the search patterns have a significant impact on the algorithm's performance, both the search speed and the search quality. The search pattern should be designed to fit the motion vector probability (MVP) distribution characteristics of the real-world sequences. In this paper, we build a directional model of MVP distribution to describe the directional-center-biased characteristic of the MVP distribution and the directional characteristics of the conditional MVP distribution more exactly based on the detailed statistical data of motion vectors of eighteen popular sequences. Three directional search patterns are firstly designed by utilizing the directional characteristics and they are the smallest search patterns among the popular ones. A new algorithm is proposed using the horizontal cross search pattern as the initial step and the horizontal/vertical diamond search pattern as the subsequent step for the fast BMME, which is called the directional cross diamond search (DCDS) algorithm. The DCDS algorithm can obtain the motion vector with fewer search points than CDS, DS or HEXBS while maintaining the similar or even better search quality. The gain on speedup of DCDS over CDS or DS can be up to 54.9%. The simulation results show that DCDS is efficient, effective and robust, and it can always give the faster search speed on different sequences than other fast block-matching algorithm in common use.

# I. INTRODUCTION

From the view point of compression in video coding, the strong temporal redundancy between the successive frames needs to be eliminated properly. Block-matching motion estimation (BMME), one of the effective and efficient methods to remove the redundancy, is pivotally adopted as the technology of inter-frame coding in the current video coding standards, such as MPEG-1 [1], MPEG-2 [2], MPEG-4 [3], H.263 [4] and the most recent H.264 [5]. The frame is partitioned into non-overlapping and equal-sized blocks (8×8) or macroblocks (16×16) in BMME, and the motion vector is obtained by finding out the displacement of the best-matched block in the reference frame (previous or past frame) within the search window centered around the original position. Matching process is performed by minimizing a certain matching criterion, and the best-matched block, where the motion vector is found, gives the minimum block-matching distortion (BMD).

However, the process of motion estimation could be the most time-consuming one in video coding and could occupy the most system resources if the full search (FS) algorithm is used to evaluate all the possible candidate blocks. In last two decades, many fast block-matching algorithms (BMA) have been proposed to accelerate the process without degrading the performance of the search algorithms fatally, such as the three-step search (TSS) algorithm [6], the new three-step search (NTSS) algorithm [7], the four-step search (4SS) algorithm [8], the block-based gradient descent search (BBGDS) algorithm [9], the diamond search (DS) algorithm [10], the unrestricted center-biased diamond search (UCBDS) algorithm [11], the hexagon-based search (HEXBS) algorithm [12], and the cross diamond search (CDS) algorithm [13], etc. In TSS, nine search points are uniformly distributed over the current search area of each step which shapes three types of square search patterns with different sizes, and the number of search points maintains 25 for each block-matching process. In consideration of the fact that the motion vectors are concentrated around the center of the search window in most cases, NTSS employs the center-biased search pattern and the halfway-stop technology, which makes it more adaptive to the motion estimation than TSS. The square search pattern is also used in 4SS and BBGDS, but the size of search pattern becomes smaller than that in TSS. 4SS does not distribute the search points uniformly but focuses on the center of the search window. The search pattern in BBGDS is the smallest square one and the number of search steps is not limited. If the motions are mainly stationary or quasi-stationary, BBGDS can give the most efficient result by checking the minimum number of search points. DS/UCBDS introduces the diamond search pattern into BMAs for the first time and the hexagon-shaped search pattern in HEXBS is another non-square one. They all try to make the search patterns approximate enough to a circle. CDS is one of the most recent fast BMAs, in which the cross search pattern is designed to fit the characteristics of the cross-center-biased motion vectors probability (MVP) distribution.

Based on the comprehensive study of MVP distribution and the relationship between the search pattern and the search result, a directional model of MVP distribution is built in this paper to describe the real-world sequences more precisely. The conditional distribution of motion vector is brought forward to show the directional characteristics of MVP distribution for the first time. A novel fast BMA called the directional cross diamond search (DCDS) algorithm is also proposed here with the horizontal cross search pattern and directional diamond search patterns. This work is improved from early versions [14, 15]. In the following section, an in-depth study on MVP distribution will be given out to build the directional model of MVP distribution. Section Ⅲ first explains the new directional cross/diamond search patterns and the DCDS algorithm, and then gives the analysis of the algorithm compared with other BMAs. Section Ⅳ presents the experiment results of the proposed algorithm and the conclusions are drawn in section Ⅴ.

## II. DIRECTIONAL MODEL OF MVP DISTRIBUTION

### A. Former Models of MVP Distribution

The search pattern with a certain shape and size has significant impact on the efficiency and the effectiveness of the search algorithm. Therefore, the search pattern is important and must be designed to fit the characteristics of MVP distribution. In fact, every discovery of the new characteristic of the MVP distribution is followed by the upgrade of the search pattern and the improvement of the search algorithm's performance.

The search pattern used in the initial fast BMAs, such as TSS, is square-shaped and the search points are distributed equally in the search window, which is based on the idea that the motion vectors are distributed with the same probability on each point in the search window. That is called the uniform MVP distribution model. In TSS, nine search points are uniformly distributed over the current search area of each step. The further study on MVP shows us that the uniform MVP distribution model can not describe the MVP distribution well and truly. In the early algorithms such as NTSS, 4SS and BBGDS, a new model called the center-biased MVP distribution model was formed gradually based on the statistic graphs of MVP distribution. Most of the motion vectors are found to be concentrative around the center of the search window no matter what the motion situation of the video sequence is, and the point with the longer distance from the center has the less probability (as depicted in Fig. 1). So the algorithms based on the center-biased model try to find out the motion vectors near the center with less block-matching computing through different methods. NTSS checks eight more proximate points around the center in the first step and employs the half-way stop technology to accelerate the search process; 4SS and BBGDS use the smaller square search patterns and half-way technology to speed up the process of searching for the motion vector near the center. Several more accurate MVP distribution models are built upon the detailed statistical data and the analyses of the motion vector in DS, HEXBS and CDS. In DS, the statistical data indicate that more than 50% of the motion vectors are enclosed in a circular support with radium of 2 pixels and centered on the position of zero motion, even more than 98% for the stationary sequence [10]. In CDS, the analyses confirm that the motion vectors concentrate on some special area around the center with different ratios: about 81.80% of the motion vectors are found located in the central 5×5 square area, moreover 77.52% and 74.76% are found in central 5×5 diamond area and cross area respectively. Then three new models were constructed in [13]: square-center-biased MVP distribution model, diamond-center-biased MVP distribution model and cross-center-biased MVP distribution model; and some new search patterns other than the square-shaped were designed. The cross-center-biased model describes the characteristics of the MVP distribution more exactly, so the CDS algorithm based on it outperforms other fast BMAs.

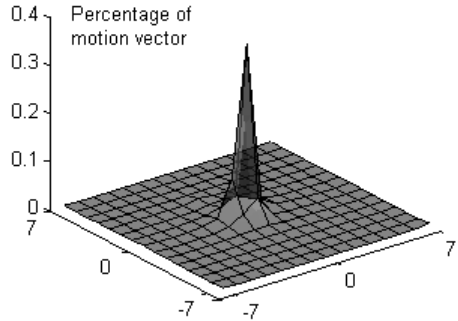

Fig. 1. The motion vector probability distribution of the "Football" sequence.

All these models are built on the assumption that the MVP is distributed regularly, equally and uniformly in different parts of the search window, and the distribution regularity can be presented by the same function. The uniform MVP distribution model hypothesizes that the MVP is the same not only in each direction but also on each position in the search window; the square-center-biased model deems that the MVP distribution is the same in eight directions (two horizontal, two vertical and four diagonal directions); the cross-center-biased model describes the MVP distribution regularity more accurately for it is same only in four directions (two horizontal and two vertical directions). The CDS fast algorithm which is based on the cross-center-biased model has the best performance. However, after some more in-depth studies on the statistical data of MVP of 18 common standard video sequences, we can see that the cross-center-biased model is not the most proper or all-around way to reflect the essence of the MVP distribution because of the existence of the directional differences.

B. Possibility of the Directional Model

There are three reasons to explain why the directional differences exist in the MVP distribution. First, the existence of the directional differences of the MVP distribution can be perceived intuitively. In a video sequence whose content is about the real world, there are two main sources of motion: one is the motion of video objects (men, cars, etc.) and the other is the movement of the camera. The motion of video objects converges in the horizontal, such as people walking and car moving, while the movements of the camera are mainly the translation for tracking the moving objects and the panning which is also in the horizontal direction. Thus it is more likely that the motion vectors could converge in the horizontal than in the vertical.

Second, the searching process of the motion vectors drops a hint of the existence of directional differences. While searching, the second step is performed on the basis of the first step, that is to say every single step always directs the search direction of the latter steps. The MVP distribution to be considered in the latter search steps should be a kind of conditional probability distribution. The conditional MVP distribution cannot be supposed to distribute equally in each direction, but should be concentrated on the direction from the search center to the current best-matched point. Consequently the conditional MVP distribution ought to have some directional differences. If the fast BMA adopts such more substantial characteristics, it should show the better performance.

Third, the existence of the directional differences can be demonstrated by the HEXBS algorithm. We call the hexagonal search pattern used in [12] the horizontal hexagonal search pattern (Fig. 2a). Obviously, there should exist another type of hexagonal search pattern, the vertical hexagonal search pattern (Fig. 2b), which can also be used in HEXBS. The fast BMAs using the horizontal and vertical hexagonal search patterns are named h-HEXBS and v-HEXBS here respectively. Their performances are compared in six aspects: 1) mean absolute distortion (MAD); 2) mean square error (MSE); 3) the number of search points (NSP) for each block; 4) peak signal noise ratio (PSNR) of the compensated frame; 5) the distance from the true motion vectors (MVs) (obtained by FS); and 6) the probability of finding the true MVs (they are all the average values over all frames in the sequence) for a "salesman" sequence (352×288, 449 frames) and a "garden" sequence (352×240, 115 frames). The simulation results are listed in Table I. The performance of h-HEXBS is almost better than that of v-HEXBS in every aspect, especially the "garden" sequence with large motion. The outcome of the different performances between h-HEXBS and v-HEXBS is only because that the directional differences exist in the real-world video sequences and the horizontal hexagonal search pattern can fit the directional characteristics of MVP distribution better.

TABLE I

SIMULATION RESULTS OF H-HEXBS AND V-HEXBS

|  | MAD | MSE | NSP | PSNR (dB) | Distance | Probability (%) |
|---|---|---|---|---|---|---|
| Salesman sequence | | | | | | |
| h-HEXBS | 2.8080 | 19.503 | 11.299 | 35.388 | 0.18036 | 94.147 |
| v-HEXBS | 2.8188 | 19.770 | 11.414 | 35.326 | 0.25322 | 90.933 |
| Garden sequences | | | | | | |
| h-HEXBS | 8.9775 | 250.86 | 13.854 | 24.303 | 0.45656 | 80.530 |
| v-HEXBS | 11.484 | 419.93 | 13.356 | 21.936 | 0.78256 | 58.701 |

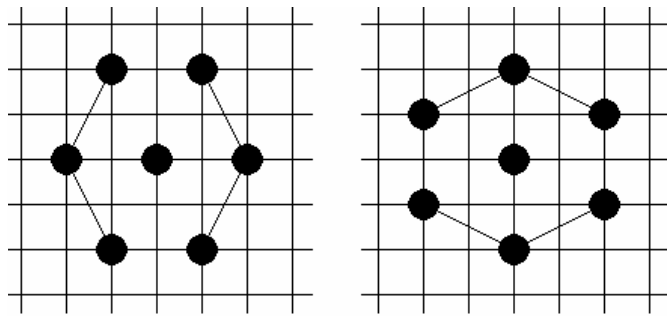

(a)          (b)

Fig. 2. The horizontal hexagonal search pattern (a) and the vertical hexagonal search pattern (b).

## C. Characteristics of the Directional Model of the MVP Distribution

For an in-depth study on the characteristics of the MVP distribution, eighteen well-known video sequences with different motion speeds/scopes, formats and lengths are computed to obtain the true MVs by FS. The low-motion content with the stationary or quasi-stationary background can be found in the video-conferencing sequences, such as "Akiyo", "Salesman", "News"; in sequences "Coastguard" and "Garden", the background changes fast in the certain direction; fast motion content and the camera's zooming and panning can be seen in the "Tennis", "Football" sequences and some other sequences have the various motion extents. MAD is employed as the block distortion measure (BDM) with block size of 16×16 and the search window size (±7) centered at the current point.

1) Directional characteristic of the MVP distribution

The statistical results of the MVP distribution are tabulated in Table II and III. MVPs accumulated at the corresponding positions of the one-quarter search window are shown as the 2-D accumulative distribution in Table II. Four types of 1-D statistics are shown in Table III: the MVP distributions of all the motion vectors accumulated in the vertical and horizontal directions ($Ax(d)$ and $Ay(d)$), and the MVP distributions in the horizontal and vertical directions ($Bx(d)$ and $By(d)$).

TABLE II

AVERAGE MVP 2-D DISTRIBUTION IN A QUARTER OF SEARCH WINDOW ±7

| Radius $\|d\|$ (vert.) | Radius $\|d\|$ (hor.) | | | | | | | |
|---|---|---|---|---|---|---|---|---|
| | 0 | 1 | 2 | 3 | 4 | 5 | 6 | 7 |
| 0 | 0.5805 | 0.1280 | 0.0591 | 0.0170 | 0.0072 | 0.0054 | 0.0026 | 0.0076 |
| 1 | 0.0572 | 0.0242 | 0.0092 | 0.0051 | 0.0041 | 0.0029 | 0.0020 | 0.0049 |
| 2 | 0.0067 | 0.0062 | 0.0034 | 0.0031 | 0.0017 | 0.0011 | 0.0010 | 0.0027 |
| 3 | 0.0031 | 0.0029 | 0.0019 | 0.0022 | 0.0012 | 0.0009 | 0.0008 | 0.0021 |
| 4 | 0.0022 | 0.0018 | 0.0014 | 0.0012 | 0.0010 | 0.0006 | 0.0007 | 0.0018 |
| 5 | 0.0012 | 0.0016 | 0.0009 | 0.0011 | 0.0007 | 0.0005 | 0.0005 | 0.0018 |
| 6 | 0.0024 | 0.0012 | 0.0008 | 0.0009 | 0.0010 | 0.0005 | 0.0007 | 0.0016 |
| 7 | 0.0015 | 0.0014 | 0.0010 | 0.0016 | 0.0012 | 0.0011 | 0.0012 | 0.0052 |

TABLE III

AVERAGE MVP 1-D DISTRIBUTION

| The distance (*d*) from the center | -7 | -6 | -5 | -4 | -3 | -2 | -1 | 0 | 1 | 2 | 3 | 4 | 5 | 6 | 7 |
|---|---|---|---|---|---|---|---|---|---|---|---|---|---|---|---|
| *Ax(d)* | 0.0117 | 0.0042 | 0.0060 | 0.0083 | 0.0114 | 0.0241 | 0.0909 | 0.6547 | 0.0762 | 0.0536 | 0.0208 | 0.0099 | 0.0069 | 0.0052 | 0.0160 |
| *Ay(d)* | 0.0065 | 0.0049 | 0.0035 | 0.0048 | 0.0070 | 0.0122 | 0.0549 | 0.8074 | 0.0548 | 0.0137 | 0.0082 | 0.0058 | 0.0048 | 0.0039 | 0.0078 |
| *Bx(d)* | 0.0042 | 0.0013 | 0.0028 | 0.0037 | 0.0049 | 0.0151 | 0.0718 | 0.5805 | 0.0562 | 0.0440 | 0.0121 | 0.0035 | 0.0026 | 0.0013 | 0.0034 |
| *By(d)* | 0.0010 | 0.0015 | 0.0006 | 0.0011 | 0.0016 | 0.0031 | 0.0295 | 0.5805 | 0.0277 | 0.0035 | 0.0015 | 0.0011 | 0.0006 | 0.0008 | 0.0006 |

As can be seen in Table II, 22.69% of the MVs are located in the horizontal direction whereas only 7.42% are located in the vertical direction (the center point is excluded). The probability of MVs in the horizontal is much more than that in the vertical with the same displacement from the central point, for example $Ax(d) > Ay(d)$ and $Bx(d) > By(d)$ ($d \neq 0$). If the MVP distribution is assumed to be the 2-D normal distribution approximately, $Ax(d)$ and $Ay(d)$ should be the 1-D normal distribution. Then the approximated normal distribution functions of the data $Ax(d)$ and $Ay(d)$ are denoted by $f_x(d)$ and $f_y(d)$,

$$f_x(d) = N(0.06666, 0.20562), \qquad (1)$$

$$f_y(d) = N(0.06668, 0.16492). \qquad (2)$$

$N(\mu, \sigma)$ is the normal function with the average $\mu$ and variance $\sigma$. One variance is about 25% bigger than the other, which means the distribution of MVP in horizontal is quite different from that in vertical.

Such directional-center-biased characteristic of the MVP distribution can also be found out from the regional statistical result of MVP. In Fig. 3, we use P, $P_\diamond$ and $P_+$ as the notations of the accumulative probabilities of central 5×5 square, diamond and cross region, use $P_\square$ and $P_\Diamond$ as the notations of that of central 3×3 square region and 5×3 flat diamond region, and they are listed below,

P = 0.8745, $P_\diamond$ = 0.8557, $P_+$ = 0.8315, $P_\square$ = 0.7899, $P_\Diamond$ = 0.8248

and   $P_\Diamond/P$ = 94.32%   $P_\Diamond/P_\diamond$ = 96.39%   $P_\Diamond/P_+$ = 99.19%

The ratios of $P_\Diamond$ to other probabilities are all very high. All the statistical results show that the MVP distribution of real-world sequences is not only center-biased or cross-center-biased, but also directional center-biased (or horizontal-center-biased (HCB), to be more exact).

| 0.0008 | 0.0013 | 0.0031 | 0.0014 | 0.0008 |
|---|---|---|---|---|
| 0.0025 | 0.0056 | 0.0295 | 0.0048 | 0.0021 |
| 0.0151 | 0.0718 | 0.5805 | 0.0562 | 0.0440 |
| 0.0024 | 0.0068 | 0.0277 | 0.0070 | 0.0021 |
| 0.0008 | 0.0018 | 0.0035 | 0.0017 | 0.0010 |

Fig. 3.   The regional statistical results of MVP.

2) Directional characteristics of the conditional MVP distribution

The MVP distribution that we focus on in the intermediate search steps should be the conditional MVP distribution because we will determine the next search direction on condition of the former search results. There are two conditional MVP distributions, the prior probability distribution and the posterior probability distribution, and they both have the directional characteristics.

The prior probability distribution of MVP is defined as the probability distribution of the global best-matched point (BMP, it is the position of the corresponding motion vector) in the search window on condition that the current BMP has been found. Let T denote the set of all the points in the search window and S denote the set of all the points covered by the search pattern in the former search steps,

$$S = \{(x, y) \mid BDM(x, y) \text{ has been calculated}, (x, y) \in T\},$$

where $BDM(x, y)$ is the block distortion measure function between the current block and the reference block whose displacement from the center is $(x, y)$. The current BMP, $P(x_p, y_p)$, is the point with the minimum distortion in S:

$$BDM(P(x_p, y_p)) = \min \{ BDM(x, y) \mid (x, y) \in S \}. \quad (3)$$

If $n_1(x, y)$ gives the number of blocks whose motion vector is on the position $(x, y)$, the prior probability of MVP distribution $D_1$ can be formularized as follows:

$$D_1(T \mid P(x_p, y_p)) = \{ n_1(x, y) \mid P = P(x_p, y_p), (x_p, y_p) \in S, (x, y) \in T \}. \quad (4)$$

To simplify the demonstration, we only give two examples of the prior probability MVP distribution based on eighteen sequences on condition that S equals the set of search point in the first step of CDS (S = {(-2, 0), (-1, 0), (0, 0), (1, 0), (2, 0), (0, 2), (0, 1), (0, -1), (0, -2)}). $D_1(T \mid P(x_p, y_p) = (2, 0))$ and $D_1(T \mid P(x_p, y_p) = (0, -2))$ are listed in Table IV. They both have the directional property that the motion vectors highly centralize in the direction from the current BMP to the center, which is closely related to the position of $P(x_p, y_p)$.

TABLE IV

THE PRIOR MOTION VECTORS DISTRIBUTION

a) $D_1(T \mid P(x_p, y_p) = (2, 0))$

| Radius $d$ (vert.) | Radius $d$ (hor.) | | | | | | | | | |
|---|---|---|---|---|---|---|---|---|---|---|
| | -2 | -1 | 0 | 1 | 2 | 3 | 4 | 5 | 6 | 7 |
| -4 | 0.0001 | 0.0001 | 0.0002 | 0.0005 | 0.0006 | 0.0005 | 0.0004 | 0.0005 | 0.0005 | 0.0020 |
| -3 | 0.0001 | 0.0001 | 0.0002 | 0.0004 | 0.0010 | 0.0011 | 0.0009 | 0.0008 | 0.0008 | 0.0024 |
| -2 | 0.0003 | 0.0002 | 0.0000 | 0.0010 | 0.0034 | 0.0034 | 0.0019 | 0.0013 | 0.0010 | 0.0031 |
| -1 | 0.0002 | 0.0005 | 0.0000 | 0.0031 | 0.0140 | 0.0098 | 0.0073 | 0.0093 | 0.0029 | 0.0057 |
| 0 | 0.0000 | 0.0000 | 0.0000 | 0.0000 | 0.5328 | 0.1329 | 0.0325 | 0.0214 | 0.0086 | 0.0217 |
| 1 | 0.0008 | 0.0010 | 0.0000 | 0.0027 | 0.0123 | 0.0069 | 0.0039 | 0.0030 | 0.0026 | 0.0071 |
| 2 | 0.0001 | 0.0002 | 0.0000 | 0.0011 | 0.0030 | 0.0034 | 0.0018 | 0.0013 | 0.0013 | 0.0046 |
| 3 | 0.0001 | 0.0001 | 0.0002 | 0.0007 | 0.0024 | 0.0051 | 0.0012 | 0.0009 | 0.0010 | 0.0038 |
| 4 | 0.0002 | 0.0001 | 0.0002 | 0.0008 | 0.0020 | 0.0017 | 0.0007 | 0.0006 | 0.0008 | 0.0032 |

b) $D_1(T \mid P(x_p, y_p) = (0, -2))$

| Radius $d$ (vert.) | Radius $d$ (hor.) | | | | | | | | |
|---|---|---|---|---|---|---|---|---|---|
| | -4 | -3 | -2 | -1 | 0 | 1 | 2 | 3 | 4 |
| -7 | 0.0034 | 0.0055 | 0.0052 | 0.0062 | 0.0222 | 0.0065 | 0.0048 | 0.0051 | 0.0042 |
| -6 | 0.0026 | 0.0019 | 0.0021 | 0.0041 | 0.0092 | 0.0128 | 0.0036 | 0.0021 | 0.0026 |
| -5 | 0.0018 | 0.0025 | 0.0024 | 0.0048 | 0.0082 | 0.0040 | 0.0025 | 0.0033 | 0.0022 |
| -4 | 0.0028 | 0.0031 | 0.0061 | 0.0079 | 0.0238 | 0.0095 | 0.0047 | 0.0035 | 0.0030 |
| -3 | 0.0034 | 0.0052 | 0.0071 | 0.0159 | 0.0549 | 0.0207 | 0.0112 | 0.0069 | 0.0043 |
| -2 | 0.0050 | 0.0096 | 0.0118 | 0.0294 | 0.1557 | 0.0310 | 0.0097 | 0.0085 | 0.0054 |
| -1 | 0.0038 | 0.0039 | 0.0044 | 0.0083 | 0.0000 | 0.0080 | 0.0066 | 0.0128 | 0.0092 |
| 0 | 0.0011 | 0.0021 | 0.0000 | 0.0000 | 0.0000 | 0.0000 | 0.0000 | 0.0043 | 0.0052 |
| 1 | 0.0007 | 0.0010 | 0.0003 | 0.0013 | 0.0000 | 0.0013 | 0.0006 | 0.0010 | 0.0012 |
| 2 | 0.0007 | 0.0010 | 0.0005 | 0.0011 | 0.0000 | 0.0013 | 0.0005 | 0.0010 | 0.0005 |

The posterior probability distribution of MVP is defined as the probability distribution of the current BMP on condition that the global BMP has been known. T and S have the same definition, and the global BMP, $Q(x_q, y_q)$, is the point with the minimum distortion in T,

$$BDM(Q(x_q, y_q),) = \min \{ BDM(x, y) \mid (x, y) \in T \}. \quad (5)$$

If $n_2(x, y)$ gives the number of blocks whose current BMP is on the position $(x, y)$, the posterior probability of MVP distribution $D_2$ can be formularized as follows:

$$D_2(S \mid Q(x_q, y_q)) = \{ n_2(x, y) \mid Q = Q(x_q, y_q), (x_q, y_q) \in T, (x, y) \in S \}. \quad (6)$$

We define S = {(x, y) | -2⩽x⩽2, -2⩽y⩽2} and give four examples of the posterior probability MVP distribution based on eighteen sequences' statistical data in Table V: $D_2(S \mid Q(x_q, y_q) = (5, 0))$, $D_2(S \mid Q(x_q, y_q) = (0, 5))$, $D_2(S \mid Q(x_q, y_q) \in T_1$, $T_1 = \{(x, y) \mid 4 \leqslant x \leqslant 7, 4 \leqslant y \leqslant 7\})$ and $D_2(S \mid Q(x_q, y_q) \in T_2$, $T_2 = \{(x, y) \mid -7 \leqslant x \leqslant -4, 4 \leqslant y \leqslant 7\})$. They all have the directional property that the current BMPs highly centralize in the direction from the global BMP to the center, which is closely related to the position of $Q(x_q, y_q)$.

TABLE V

THE POSTERIOR MOTION VECTORS DISTRIBUTION

a) $D_2(S \mid Q(x_q, y_q) = (5, 0))$

| Radius d (vert.) | Radius d (hor.) | | | | |
|---|---|---|---|---|---|
| | -2 | -1 | 0 | 1 | 2 |
| 2 | 0.0026 | 0.0017 | 0.0028 | 0.0080 | 0.0499 |
| 1 | 0.0032 | 0.0028 | 0.0039 | 0.0047 | 0.0391 |
| 0 | 0.0176 | 0.0170 | 0.0454 | 0.0312 | 0.3129 |
| -1 | 0.0039 | 0.0024 | 0.0075 | 0.0054 | 0.0441 |
| -2 | 0.0022 | 0.0015 | 0.0030 | 0.0026 | 0.0275 |

b) $D_2(S \mid Q(x_q, y_q) = (0, 5))$

| Radius d (vert.) | Radius d (hor.) | | | | |
|---|---|---|---|---|---|
| | -2 | -1 | 0 | 1 | 2 |
| 2 | 0.033 | 0.033 | 0.127 | 0.034 | 0.034 |
| 1 | 0.006 | 0.021 | 0.115 | 0.019 | 0.006 |
| 0 | 0.007 | 0.02 | 0.124 | 0.027 | 0.008 |
| -1 | 0.006 | 0.013 | 0.063 | 0.009 | 0.002 |
| -2 | 0.003 | 0.012 | 0.038 | 0.01 | 0.004 |

c) $D_2(S \mid Q(x_q, y_q) \in T_1)$

| Radius d (vert.) | Radius d (hor.) | | | | |
|---|---|---|---|---|---|
| | -2 | -1 | 0 | 1 | 2 |
| 2 | 0.0318 | 0.0152 | 0.0240 | 0.0206 | 0.2288 |
| 1 | 0.0077 | 0.0080 | 0.0328 | 0.0090 | 0.0239 |
| 0 | 0.0074 | 0.0113 | 0.0359 | 0.0156 | 0.0220 |
| -1 | 0.0068 | 0.0059 | 0.0236 | 0.0070 | 0.0142 |
| -2 | 0.0290 | 0.0062 | 0.0128 | 0.0073 | 0.0338 |

d) $D_2(S \mid Q(x_q, y_q) \in T_2)$

| Radius $d$ (vert.) | Radius $d$ (hor.) | | | | |
|---|---|---|---|---|---|
| | -2 | -1 | 0 | 1 | 2 |
| 2 | 0.2103 | 0.0320 | 0.0346 | 0.0166 | 0.0462 |
| 1 | 0.0352 | 0.0157 | 0.0358 | 0.0111 | 0.0148 |
| 0 | 0.0297 | 0.0280 | 0.0549 | 0.0175 | 0.0138 |
| -1 | 0.0192 | 0.0098 | 0.0288 | 0.0088 | 0.0088 |
| -2 | 0.0318 | 0.0114 | 0.0163 | 0.0095 | 0.0202 |

## D. Directional Model of the MVP Distribution

The directional model of the MVP distribution can be built easily based on the former analyses: the motion vector distribution is not equal or same in the different directions, but is horizontal-center-biased. The MVPs concentrate more heavily in the horizontal directions than in the vertical. The conditional distribution of MVP has the directional properties so that the direction from the center to the current BMP gives the rough orientation of the subsequent search. These two characteristics will help improve the performance of the first and latter search steps in fast BMA.

## III. DIRECTIONAL CROSS DIAMOND SEARCH ALGORITHM

### A. Search Patterns

The search patterns in the previous BMAs are symmetrical in all four horizontal and vertical directions, which do not correspond with the directional characteristics of the MVP distribution. Therefore, a new kind of search pattern needs to be designed to find the motion vector more quickly and directly in the proper direction. Based on the horizontal center-biased MVP distribution and directional characteristics of the conditional distribution of MVP proposed above, the horizontal cross search pattern (HCSP) and directional diamond search patterns (DDSP), as depicted in Fig. 4, are proposed in the new BMA, which is termed the directional cross diamond search (DCDS) algorithm.

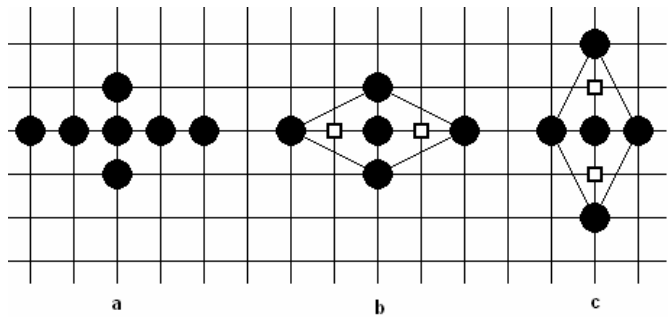

Fig. 4. Three directional search patterns, a) HCSP, b) HDSP, c)VDSP.

HCSP is proposed according to the horizontal center-biased MVP distribution. It has only seven points, the minimum number of search points compared to other patterns, which makes DCDS more efficient in the first step than other BMAs. The search efficiency of the first step ($\eta_1$) is defined as the average probability on every search point in the first step,

$$\eta_1 = P_1 / n_1, \qquad (7)$$

where $P_1$ is the sum of the probability in the area covered by the search pattern and $n_1$ is the number of search points of the search pattern. We use the data in Table II to calculate $\eta_1$ of different patterns and the results are tabulated in Table VI. HCSP has the highest $\eta_1$ among all the current used patterns.

TABLE VI

THE SEARCH EFFICIENCY OF THE FIRST STEP OF DIFFERENT BMAS

| | |
|---|---|
| 3×3 square search pattern in BBGDS | $\eta_1^S = P_1^S / n_1^S = 0.7899/9 = 0.0878$ |
| 5×5 diamond search pattern in DS | $\eta_1^D = P_1^D / n_1^D = 0.6705/9 = 0.0745$ |
| 5×5 cross search pattern in CDS | $\eta_1^C = P_1^C / n_1^C = 0.8315/9 = 0.0924$ |
| 5×5 hexagonal search pattern in HEXBS | $\eta_1^H = P_1^H / n_1^H = 0.6457/7 = 0.0922$ |
| Horizontal cross search pattern in DCDS | $\eta_1^{HCSP} = P_1^{HCSP} / n_1^{HCSP} = 0.8248/7 = 0.1178$ |

Two kinds of configuration of directional diamond search patterns are designed to fit the directional characteristic of the conditional distribution of MVP: one is called the horizontal diamond search pattern (HDSP); the other is called the vertical diamond search pattern (VDSP). They are the smallest search patterns with the directional property, only 5 points in each pattern. They will be more effective to reach the BMP in the most direct and shortest way if they are used in the intermediate search steps by checking the minimum points each step. In these patterns, the points with the distance 2 to the center point are called the distant points and the points with the distance 1 are called the near points; the part of the pattern in the direction where the distant points are located is called the long wing and the other part is called the short wing; the points with the distance 1 to the center point on the long wings are called the middle points (the hollow squares in Fig. 4). The middle points play an important role in the search process because of their shortest distance to the center, although they do not belong to the HDSP or VDSP. All these three search patterns are used in the DCDS algorithm but the different steps.

## B. DCDS Algorithm

The DCDS algorithm is quite different from any other fast BMAs in: 1) the search patterns used in DCDS have the minimum number of points; 2) the directional search patterns are used; and 3) the switching strategy of the different search patterns in the middle steps is adopted necessarily. DCDS exploits the characteristics of the directional model of MVP distribution completely, replacing the cross search pattern with the HCSP in the first step and the diamond search pattern with HDSP/VDSP compared to CDS. Below summarizes the DCDS algorithm.

Step1: HCSP is centered at the origin of the search window and set as the current search pattern (CSP). If the current BMP occurs at the center of the CSP, the search process stops and the motion vector is found on the center; otherwise, go to step2;

Step2: Update the CSP to HDSP or VDSP according to the switching strategy one, put the center of the CSP on the current BMP, and calculate distortion measure to find the new current BMP. If the current BMP occurs at the center point, go to step4; otherwise go to step3;

Step3: Update the CSP according to the switching strategy two, put the center of the CSP on the current BMP, and calculate distortion measure to find the new current BMP. If the current BMP occurs at the center point, go to step4; otherwise repeat this step continuously;

Step4: Identify the BMP from the two middle points and center point of the CSP. The position of the global BMP will be the final solution of the motion vector.

There are two switching strategies in DCDS algorithm, in which the next search pattern is determined by the location of the current BMP in the CSP. They can be described as follows (Fig. 5):

Switching strategy one: If the BMP is in the horizontal direction, the CSP will be switched to HDSP, otherwise switched to VDSP.

Switching strategy two: If the BMP occurs on the near point, the CSP is switched to the other DDSP (HDSP or VDSP); otherwise, keep using the current search pattern.

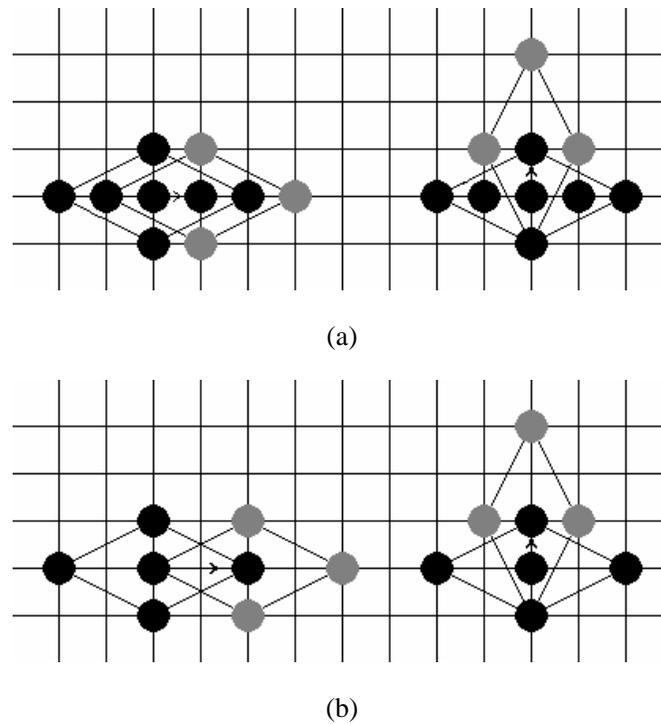

(a)

(b)

Fig. 5.  Switching strategy one (a) and two (b).

From the procedure, the total number of search points per block can be derived easily,

$$N_{DCDS}(m_x, m_y) = \begin{cases} 7 & \text{If the motion vector is zero} \\ 7 + 3 \times n + 1 & \text{If CSP changes in the last step} \\ 7 + 3 \times n + 2 & \text{If CSP maintains in the last step} \end{cases}, \quad (8)$$

where $(m_x, m_y)$ is the final motion vector found, and $n$ is the number of implementation of intermediate steps (Step 2 or Step 3).

A complete search procedure of DCDS is given in Fig. 6.

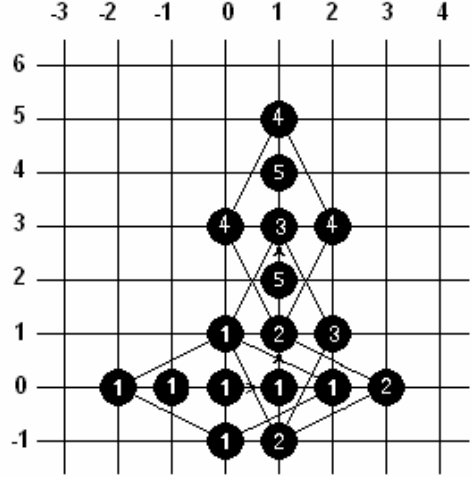

Fig. 6. Example of DCDS, motion vector is located on (1, 3).

C. Analysis of the Proposed DCDS Algorithm

1) The half-way stop and the pattern-switching technique

The half-way stop technique is also used in DCDS as in NTSS, DS, and CDS, etc. If the motion vector is zero, the DCDS algorithm will stop at the first step, and only seven points need to be checked, which is less than any other fast BMAs. If the small motion vector is (±1, 0) or (0, ±1), the search process will only check 10 or 11 points respectively, which is also the minimum number. Generally, 58.05% and 18.32% of the search procedure will stop after using one and two search patterns and lead to the ideal speedup of 32 and 21 times as compared to FS by exploiting the data in Table II. But the pattern-switching technique with two switching strategies is first introduced into the intermediate steps of the fast BMAs by DCDS when using HDSP and VDSP alternatively. In the former fast BMAs, the search pattern is only switched to another different one after the first step or in the last step, while in DCDS the pattern-switching can happen in every step without much extra computational load. That is determined by the guideline of DCDS which tries to make the search pattern fit the directional characteristic of the current MVP distribution. Although three different patterns are used together in DCDS and the patterns are not symmetrical, the number of new points to be checked in the next step is always three (the gray points in Fig. 5.) as a big contrast to the symmetrical square or diamond search pattern in BBGDS, DS or CDS, which need check three or five points in different search directions.

2) The number of search points on the ideal condition

The uni-modal error surface assumption of the BDM is one ideal condition of the MVP distribution: the BDM of matching blocks increases monotonically away from the global minimum distortion. It produces us an identical condition to evaluate the general performance of different algorithms though it is seldom right to reflect the actual distribution. We set up such an ideal condition: the distortion between the current block $P_0$ and the best-matched reference block $MV(x_{mv}, y_{mv})$ is zero, and the block-matching distortion of other reference block $P(x_p, y_p)$ is defined as its Euclid distance to the best-matched block, and then calculate the number of search points on each position of the search window (as listed in Table VII).

TABLE VII

THE NUMBER OF SEARCH POINTS OF DIFFERENT FAST BMAS ON AN IDEAL CONDITION

| NTSS | 0 | 1 | 2 | 3 | 4 | 5 | 6 | 7 | DS | 0 | 1 | 2 | 3 | 4 | 5 | 6 | 7 |
|---|---|---|---|---|---|---|---|---|---|---|---|---|---|---|---|---|---|
| 0 | 17 | 20 | 20 | 33 | 33 | 33 | 33 | 33 | 0 | 13 | 13 | 18 | 18 | 23 | 23 | 27 | 27 |
| 1 | 20 | 22 | 22 | 33 | 33 | 33 | 33 | 33 | 1 | 13 | 16 | 16 | 21 | 21 | 26 | 26 | 27 |
| 2 | 20 | 22 | 22 | 22 | 33 | 33 | 33 | 33 | 2 | 18 | 16 | 19 | 19 | 24 | 24 | 28 | 28 |
| 3 | 33 | 33 | 22 | 33 | 33 | 33 | 33 | 33 | 3 | 18 | 21 | 19 | 22 | 22 | 27 | 27 | 28 |
| 4 | 33 | 33 | 33 | 33 | 33 | 33 | 33 | 33 | 4 | 23 | 21 | 24 | 22 | 25 | 25 | 29 | 29 |
| 5 | 33 | 33 | 33 | 33 | 33 | 33 | 33 | 33 | 5 | 23 | 26 | 24 | 27 | 25 | 28 | 28 | 29 |
| 6 | 33 | 33 | 33 | 33 | 33 | 33 | 33 | 33 | 6 | 27 | 26 | 28 | 27 | 29 | 28 | 29 | 29 |
| 7 | 33 | 33 | 33 | 33 | 33 | 33 | 33 | 33 | 7 | 27 | 27 | 28 | 28 | 29 | 29 | 29 | 27 |
| CDS | 0 | 1 | 2 | 3 | 4 | 5 | 6 | 7 | DCDS | 0 | 1 | 2 | 3 | 4 | 5 | 6 | 7 |
| 0 | 9 | 11 | 19 | 19 | 25 | 25 | 29 | 29 | 0 | 7 | 10 | 11 | 11 | 15 | 15 | 17 | 17 |
| 1 | 11 | 17 | 17 | 23 | 23 | 28 | 28 | 29 | 1 | 11 | 13 | 14 | 17 | 17 | 20 | 19 | 20 |
| 2 | 19 | 17 | 22 | 22 | 26 | 26 | 30 | 30 | 2 | 11 | 16 | 14 | 20 | 17 | 23 | 19 | 23 |
| 3 | 19 | 23 | 22 | 25 | 25 | 29 | 29 | 30 | 3 | 15 | 17 | 17 | 20 | 20 | 23 | 23 | 23 |
| 4 | 25 | 23 | 26 | 25 | 28 | 28 | 31 | 31 | 4 | 15 | 20 | 23 | 23 | 26 | 26 | 28 | 26 |
| 5 | 25 | 28 | 26 | 29 | 28 | 31 | 31 | 31 | 5 | 18 | 20 | 20 | 23 | 23 | 26 | 26 | 27 |
| 6 | 29 | 28 | 30 | 29 | 31 | 31 | 32 | 32 | 6 | 18 | 22 | 25 | 25 | 28 | 28 | 30 | 29 |
| 7 | 29 | 29 | 30 | 30 | 31 | 31 | 32 | 30 | 7 | 19 | 21 | 21 | 25 | 29 | 28 | 31 | 29 |

The search window is divided into six parts ($P_1$, $P_2$, $\cdots$, $P_6$ in Fig. 7) and the sums of MVP of each part are given as following: $P_1$=0.4, $P_2$=0.2, $P_3$=0.2, $P_4$=0.1, $P_5$=0.1, $P_6$=0 (distribution 1), which is a typical center-biased distribution. We also can calculate $P_1$, ..., $P_6$ in Table II as distribution 2. The average number of search point (ANSP) all-over the search window on distribution 1 and 2 of different BMAs are listed in Table VIII. The DCDS algorithm always searches the minimum points among these algorithms.

TABLE VIII

THE AVERAGE NUMBER OF SEARCH POINTS OF DIFFERENT BMAS

| Fast BMA | TSS | NTSS | 4SS | BBGDS | DS | CDS | HEXBS | DCDS |
|---|---|---|---|---|---|---|---|---|
| Distribution 1 | 25 | 21.5 | 18.9 | 14.2 | 15.8 | 14.9 | 13.1 | 11.7 |
| Distribution 2 | 25 | 19.9 | 18.0 | 12.4 | 14.8 | 12.5 | 12.1 | 9.7 |

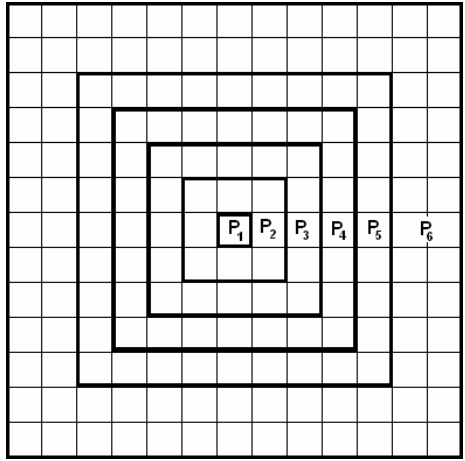

Fig. 7. Six parts of the search window ±7.

## IV. SIMULATION RESULTS

The DCDS algorithm is simulated using the luminance of 18 popular sequences which consist of different types of motion content. The MAD as the BDM, block size of 16, and search window size of ±7 is used for all BMAs. The proposed DCDS algorithm is compared against six other BMAs: FS, NTSS, 4SS, DS, HEXBS and CDS in six aspects: 1) MAD; 2) MSE; 3) NSP for each block; 4) PSNR of the compensated frame; 5) distance from the true MVs; and 6) probability of finding the true MVs.

### A. Performance Comparison of the Whole Sequence (tables)

The average performance parameters of three sequences "Salesman", "Stefan" and "Coastguard" are given in the Table IX.

### TABLE IX

### THE AVERAGE PERFORMANCE PARAMETERS OF DIFFERENT BMAS

Salesman Sequence

|       | MAD   | MSE    | NSP    | PSNR   | Distance | Probability |
|-------|-------|--------|--------|--------|----------|-------------|
| FS    | 2.751 | 18.638 | 225    | 35.570 | 0        | 1           |
| NTSS  | 2.759 | 18.850 | 17.951 | 35.529 | 0.07903  | 0.98181     |
| 4SS   | 2.789 | 19.261 | 17.371 | 35.433 | 0.19688  | 0.95681     |
| DS    | 2.785 | 19.193 | 13.795 | 35.449 | 0.20117  | 0.95605     |
| HEXBS | 2.819 | 19.770 | 11.414 | 35.326 | 0.25322  | 0.90933     |
| CDS   | 2.761 | 18.970 | 9.768  | 35.509 | 0.08401  | 0.98018     |
| DCDS  | 2.764 | 19.045 | 8.100  | 35.496 | 0.08780  | 0.97808     |

Stefan Sequence

|  | MAD | MSE | NSP | PSNR | Distance | Probability |
|---|---|---|---|---|---|---|
| FS | 10.686 | 449.91 | 255 | 23.594 | 0 | 1 |
| NTSS | 11.327 | 502.46 | 25.525 | 23.108 | 1.1193 | 0.81635 |
| 4SS | 11.704 | 534.84 | 20.678 | 22.713 | 1.4794 | 0.78867 |
| DS | 11.685 | 537.01 | 19.477 | 22.714 | 1.3362 | 0.56029 |
| HEXBS | 11.903 | 549.57 | 14.770 | 22.573 | 1.5832 | 0.73490 |
| CDS | 11.764 | 543.83 | 18.818 | 22.646 | 1.3826 | 0.80019 |
| DCDS | 11.845 | 552.00 | 12.759 | 22.604 | 1.4292 | 0.77975 |

Coastguard Sequence

|  | MAD | MSE | NSP | PSNR | Distance | Probability |
|---|---|---|---|---|---|---|
| FS | 4.611 | 60.621 | 255 | 30.602 | 0 | 1 |
| NTSS | 4.656 | 62.449 | 21.291 | 30.555 | 0.08497 | 0.97716 |
| 4SS | 4.689 | 64.754 | 19.799 | 30.479 | 0.09877 | 0.98298 |
| DS | 4.669 | 63.739 | 17.667 | 30.518 | 0.08024 | 0.98918 |
| HEXBS | 5.672 | 86.61 | 12.129 | 29.043 | 0.67065 | 0.49634 |
| CDS | 4.671 | 63.692 | 16.857 | 30.521 | 0.08082 | 0.98892 |
| DCDS | 4.687 | 64.634 | 10.885 | 30.500 | 0.09137 | 0.98609 |

When applied to stationary or quasi-stationary sequence, such as "Salesman", DCDS and CDS algorithm have the similar performance according to the PSNR of the compensated frame while the search speed (measured by the number of search point) of DCDS is 20.6% faster than that of CDS. But when applied to the sequence having large motion content and various motion directions, DCDS can speed up the search progress significantly. Take the sequence "Coastguard" as the example, the NSP of DCDS and CDS are 10.885 and 16.857 respectively, so DCDS achieves 54.9% speed-up with only 0.021dB of degradation in the quality. Other aspects of DCDS and CDS are all quite similar.

B. Performance Comparison Frame-wise (figures)

Fig. 8 and 9 illustrate the frame-by-frame comparison of PSNR and NSP after applying FS, NTSS, 4SS, DS, HEXBS, CDS and DCDS to "Salesman" and "Coastguard". They clearly demonstrate the robust and superior performance of the proposed DCDS algorithm to other BMAs in terms of the average number of search points with the similar or even better distortion error in terms of PSNR.

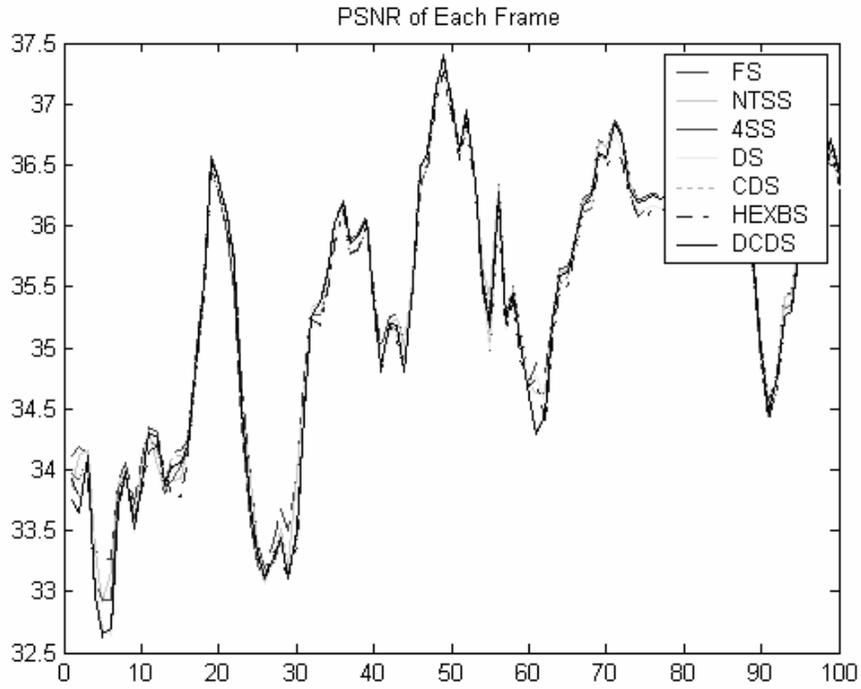

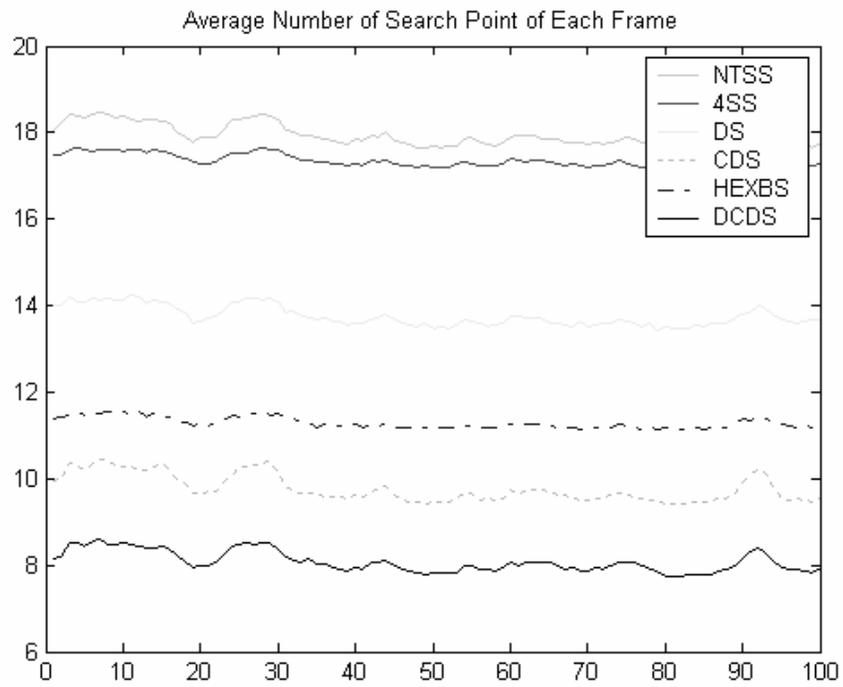

Fig. 8. Frame-wise performance comparison between different BMAs on sequence "Salesman" by (a) PSNR and (b) average NSP.

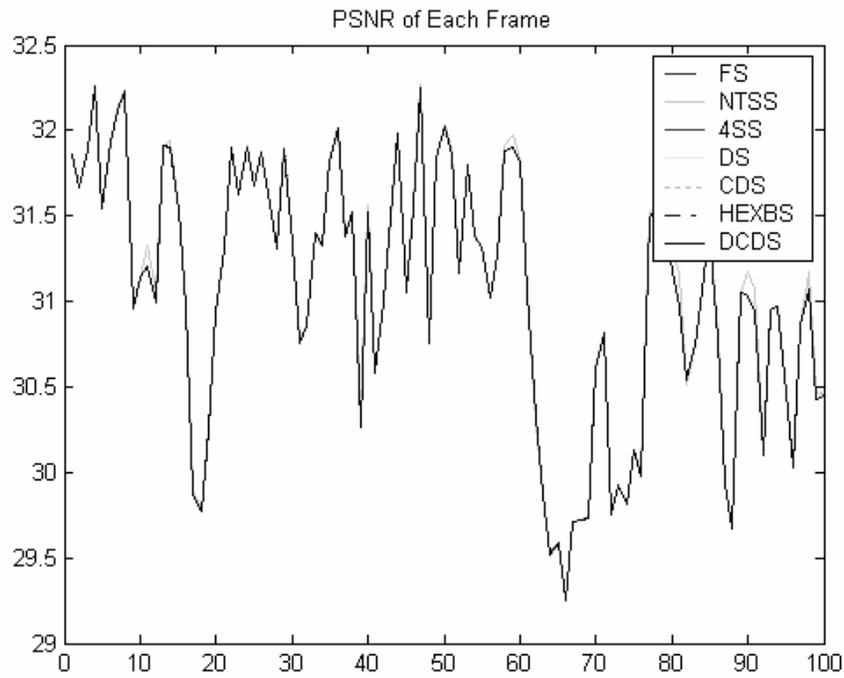

(a)

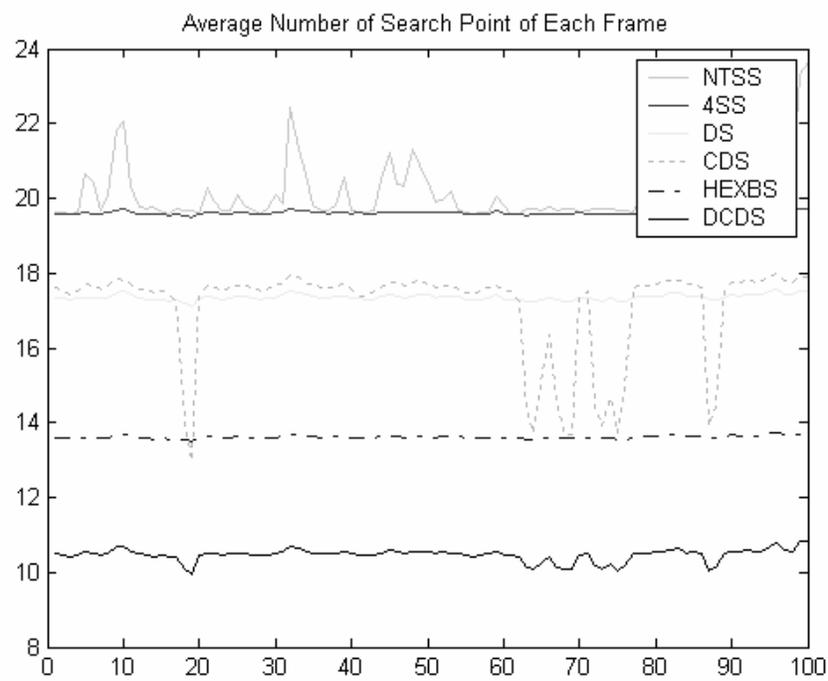

(b)

Fig. 9. Frame-wise performance comparison between different BMAs on sequence "Coastguard" by (a) PSNR and (b) average NSP.

## C. Further Discussion

In the last step of DCDS, two middle points are checked when the current BMP is located in the center of CSP. Actually, we can consider that if the distortion on one distant point is smaller, the middle point on the same side should more likely give the smaller distortion than the middle point on the other side. So if we can use the comparison information between the distortions of two distant points, one checking point will be saved probably. Of course, the trade-off of the speedup is the bigger average distortion of the search result statistically. This algorithm is called the simplified DCDS ($DCDS^S$) algorithm. $DCDS^S$ can give faster search speed. In Table X tabulating the performances between DCDS and $DCDS^S$ on sequence "Salesman", "Stefan" and "Football", the saving number of point is up to 0.361 and the degradation of the frame quality is up to 0.364dB. DCDS and $DCDS^S$ should be used in different situation according to the different requirements.

TABLE X

THE AVERAGE PERFORMANCE PARAMETERS BETWEEN DCDS AND $DCDS^S$

Salesman Sequence

|  | MAD | MSE | NSP | PSNR | Distance | Probability |
|---|---|---|---|---|---|---|
| DCDS | 2.764 | 19.045 | 8.1000 | 35.496 | 0.08780 | 0.97808 |
| $DCDS^S$ | 2.766 | 19.115 | 7.9383 | 35.482 | 0.09092 | 0.97536 |

Stefan Sequence

|  | MAD | MSE | NSP | PSNR | Distance | Probability |
|---|---|---|---|---|---|---|
| DCDS | 11.845 | 552.00 | 12.759 | 22.604 | 1.4292 | 0.77975 |
| $DCDS^S$ | 12.279 | 572.40 | 12.445 | 22.240 | 1.5469 | 0.66299 |

Coastguard Sequence

|  | MAD | MSE | NSP | PSNR | Distance | Probability |
|---|---|---|---|---|---|---|
| DCDS | 4.687 | 64.634 | 10.885 | 30.500 | 0.09137 | 0.98609 |
| $DCDS^S$ | 4.794 | 67.417 | 10.524 | 30.346 | 0.17778 | 0.89960 |

## V. CONCLUSION

The in-depth study of the fast block matching motion estimation algorithm based on search pattern is performed in this paper. The evolutionary progress of the fast BMAs based on search pattern is discovered to fit the MVP distribution. A new directional model of MVP distribution which can reflect the characteristics of MVP distribution more precisely is brought forward from the statistical results of the actual data. The conception of the conditional MVP distribution in this paper is given, including the prior probability distribution and the posterior probability distribution of motion vectors. Three new search patterns based on the directional model are designed and a new fast BMA called DCDS is proposed by using three new patterns. Both the experimental results and the theoretical analyses have proved that the new algorithm is much more effective and efficient than other algorithms.

Footnotes:

Hongjun Jia and Li Zhang are both with the Department of Electronic Engineering, Tsinghua University, Beijing 100084, China (phone: 8610-62781436; fax: 8610-62770317; e-mail: jia.22@osu.edu, chinazhangli@mail.tsinghua.edu.cn).

All figure captions

Fig. 1.   The motion vector probability distribution of the "Football" sequence.

Fig. 2.   The horizontal hexagonal search pattern and the vertical hexagonal search pattern.

Fig. 3.   The regional statistical result of MVP.

Fig. 4.   Three directional search patterns, a) HCSP, b) HDSP, c)VDSP.

Fig. 5.   Switching strategy one (a) and two (b).

Fig. 6.   Example of DCDS, motion vector is located on (1, 3).

Fig. 7.   Six parts of the search window ±7.

Fig. 8.   Frame-wise performance comparison between different BMAs on sequence "Salesman" by (a) PSNR and (b) average NSP.

Fig. 9.   Frame-wise performance comparison between different BMAs on sequence "Coastguard" by (a) PSNR and (b) average NSP.